\documentclass[runningheads]{llncs}

\usepackage{eccv}

\usepackage{eccvabbrv}

\usepackage{graphicx}
\usepackage{booktabs}

\usepackage[accsupp]{axessibility}  

\usepackage[pagebackref,breaklinks,colorlinks,citecolor=eccvblue]{hyperref}

\usepackage{orcidlink}

\begin{document}

\title{7th ABAW Competition: Multi-Task Learning and Compound Expression Recognition} 

\titlerunning{7th ABAW Competition: MTL and CER}

\author{First Author\inst{1}\orcidlink{0000-1111-2222-3333} \and
Second Author\inst{2,3}\orcidlink{1111-2222-3333-4444} \and
Third Author\inst{3}\orcidlink{2222--3333-4444-5555}
}

\author{Dimitrios Kollias\inst{1} 
\and
Stefanos Zafeiriou\inst{2} 
\and
Irene Kotsia\inst{3} 
\and
Abhinav Dhall\inst{4}
\and
Shreya Ghosh\inst{5}
\and
Chunchang Shao\inst{1}  
\and
Guanyu Hu\inst{1,6}  
}

\authorrunning{D.~Kollias et al.}

\institute{Queen Mary University of London, UK \\
\email{d.kollias@qmul.ac.uk}\\
\and
Imperial College London, UK\\
\and
Cogitat, UK \\
\and
Flinders University, Australia\\
\and
Curtin University, Australia\\
\and 
Xi'an Jiaotong University, China
}

\maketitle

\begin{abstract}
This paper describes the 7th Affective Behavior Analysis in-the-wild (ABAW)  Competition, which is part of the respective Workshop held in conjunction with ECCV 2024. The 7th ABAW Competition addresses novel challenges in understanding human expressions and behaviors, crucial for the development of human-centered technologies. The Competition comprises of two sub-challenges: i) Multi-Task Learning  (the goal is to learn at the same time, in a multi-task learning setting, to estimate two continuous affect dimensions, valence and arousal, to recognise between the mutually exclusive classes of the 7 basic expressions and 'other'), and to detect 12 Action Units); and ii) Compound Expression Recognition (the target is to recognise between the 7 mutually exclusive compound expression classes).  s-Aff-Wild2, which is a static version of the A/V Aff-Wild2 database and contains annotations for valence-arousal, expressions and Action Units, is utilized for the purposes of the Multi-Task Learning Challenge; a part of C-EXPR-DB, which is an A/V in-the-wild database with compound expression annotations, is utilized for the purposes of the Compound Expression Recognition Challenge.
In this paper, we introduce the two challenges, detailing their datasets and the protocols followed for each. We also outline the evaluation metrics, and highlight the baseline systems and their results. Additional information about the competition can be found at \url{https://affective-behavior-analysis-in-the-wild.github.io/7th}.

\end{abstract}

\section{Introduction}
\label{sec:intro}

Facial behavior analysis, bridging computer vision, physiology, and psychology, has broad applications including medicine, e-learning, marketing, entertainment and law. Recent advancements in large-scale datasets and deep learning have propelled the field forward significantly.

The 7th Affective Behavior Analysis in-the-wild (ABAW) Workshop and Competition continues to promote interdisciplinary collaboration by uniting experts from academia, industry, and government. Held alongside ECCV 2024, this workshop focuses on analyzing affective behavior in real-world contexts. This research is pivotal for advancing human-centered technologies such as human-computer interaction (HCI) systems and intelligent digital assistants. By improving our understanding of human emotions and behaviors, machines can more effectively engage with users across diverse contexts, including variations in age, gender, and social background, thereby fostering trust and improving real-life interactions.

The ABAW Competition is a crucial component of the Workshop and consists of two distinct challenges.

The first challenge focuses on Multi-Task Learning (MTL). This challenge aims to simultaneously address three tasks: (i) estimating two continuous affect dimensions (valence and arousal), (ii) recognizing eight distinct classes of facial expressions, and (iii) detecting the activation of 12 Action Units (AUs). Valence represents the positivity or negativity of an affective state on a continuous scale ranging from -1 to 1, while arousal measures the intensity of the affective state, from active to passive, also on a scale from -1 to 1. The eight expression classes include the six basic expressions (anger, disgust, fear, happiness, sadness, and surprise), the neutral state, and an 'other' category for affective states that do not fit into the other seven classes. Action Units correspond to specific facial muscle movements, with the selected AUs for this challenge being: AU1, AU2, AU4, AU6, AU7, AU10, AU12, AU15, AU23, AU24, AU25, and AU26.

For this Challenge, a static version of the Aff-Wild2 database \cite{zafeiriou2017aff,kollias2017recognition,kollias2019expression,kollias2020analysing,kollias2021analysing,kollias2021affect,kollias2021distribution,kollias2022abaw,kollias2019face,kollias2023abaww,kollias2019deep,kollias2023abaww,kollias2023multi,kollias20246th,kollias2024distribution,kolliasijcv,kollias2023deep2,kollias2020va,kollias2018photorealistic}, named s-Aff-Wild2, is used. It consists of around 222K images that  contain annotations in terms of the previously mentioned tasks.

The participants are allowed to use the provided s-Aff-Wild2 database and/or any publicly available or private databases. The participants are not allowed to use the (A/V) Aff-Wild2 database (images and annotations). Teams are allowed to use any -publicly or not- available pre-trained model (as long as it has not been pre-trained on Aff-Wild2). The pre-trained model can be pre-trained on any task (e.g., VA estimation, Expression Recognition, AU detection, Face Recognition). Any methodological solution will be accepted for this Challenge.

The second Challenge targets Compound Expression (CE) Recognition. This task involves identifying seven distinct compound expressions in each frame of the provided dataset. The compound expressions to be recognized are: Fearfully Surprised, Happily Surprised, Sadly Surprised, Disgustedly Surprised, Angrily Surprised, Sadly Fearful, and Sadly Angry.

For this Challenge, a subset of the C-EXPR-DB database \cite{kollias2023multi} is utilized, comprising 56 videos. C-EXPR-DB is an audiovisual (A/V) in-the-wild database containing a total of 400 videos and approximately 200,000 frames. The subset provided for this Challenge includes 56 videos with roughly 26,500 frames, which are unannotated.

Teams are allowed to use any -publicly or not- available pre-trained model and any -publicly or not- available database (that contains any annotations, e.g. VA, basic or compound expressions, AUs).

The seventh iteration of the ABAW Competition will take place as part of the ABAW Workshop, held in conjunction with ECCV 2024. This competition continues the successful legacy of previous ABAW Competitions, which have been held alongside IEEE CVPR 2024, 2023, 2022 and 2017, ECCV 2022, ICCV 2021 and IEEE FG 2020. These events have consistently attracted numerous teams from academia and industry worldwide \cite{zhang2023facial,zhang2023multimodal,wang2023spatio,savchenko2023emotieffnet,vu2023vision,wang2023facial,yin2023multi,zou2023spatial,kollias2023facernet,zhou2023continuous,zhang2023facial,savchenko2023emotieffnet,xue2023exploring,yu2023exploring,gera2023abaw,nguyen2023transformer,mutlu2023tempt,zou2023spatial,shu2023mutilmodal,kim2023multi,deng2020multitask,li2021technical,zhang2020m,do2020affective,chen2017multimodal,weichi,deng2021towards,zhang2021prior,vu2021multitask,wang2021multi,zhang2021audio,xie2021technical,jin2021multi,antoniadis2021audiovisual,oh2021causal,kuhnke2020two,gera2020affect,dresvyanskiy2020audio,youoku2020multi,liu2020emotion,gera2021affect,mao2021spatial,pahl2020multi,ji2020multi,han2016incremental,deng2020fau,saito2021action,meng2022multi,zhang2022continuous,nguyen2022ensemble,savchenko2022frame,karas2022continuous,rajasekar2022joint,zhang2022transformer,yu2022multi,kim2022facial,phan2022expression,xue2022coarse,jeong2022facial,wang2022facial,hoai2022attention,tallec2022multi,wang2022multi,jiang2022facial,deng2022multiple,haider2022ensemble,sun2022two,chang2022multi,gera2022ss,li2022facial,wang2022hybrid,nguyen2022multi,savchenko2022hse,li2022affective,zhang2022emotion,lee2022byel,jeong2022learning,lei2022mid,kollias2024distribution,min2024emotion,wen2024multimodal,waligora2024joint,kim2024cca,savchenko2024hsemotion,yu2024multimodal,dresvyanskiy2024sun,zhou2024boosting,praveen2024recursive,zhang2024affective,zhou2024boosting,yu2024exploring,lin2024robust,nguyen2024emotic,wen2024multimodal,yu2024aud,ryumina2024audio,wang2024zero,hallmen2024unimodal,kollias2023facernet,hu2024bridging,psaroudakis2022mixaugment}.

\section{Competition Corpora}\label{corpora}

Below, we provide an overview of the databases used in each Challenge. For detailed information, we recommend referring to the original documentation. Additionally, we describe the pre-processing steps taken for the Challenges, which involve cropping and aligning all image frames. These steps were utilized in our baseline experiments.

\subsection{Multi-Task Learning Challenge}

A static version of the Aff-Wild2 database, termed s-Aff-Wild2, was created by selecting specific frames from the original database. This Challenge's database includes 221,928 images annotated for: (i) valence and arousal; (ii) eight expressions (six basic expressions: anger, disgust, fear, happiness, sadness, and surprise; the neutral state; and an 'other' category for all other affective states); and (iii) twelve Action Units (AUs).

\begin{figure}[!h]
\centering
\includegraphics[height=6.5cm]{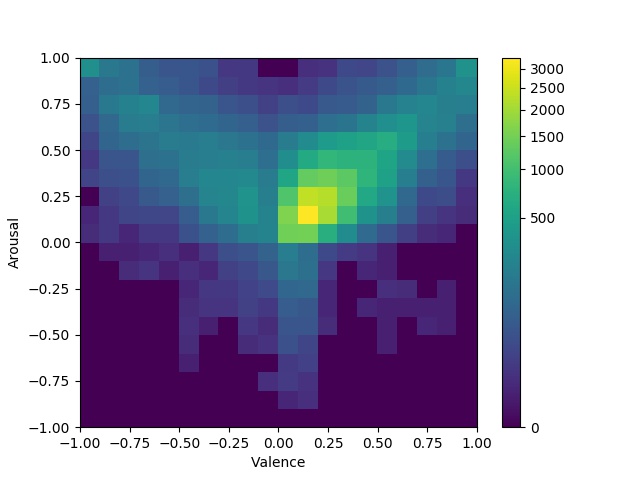}
\caption{Multi-Task-Learning Challenge: 2D Valence-Arousal Histogram of Annotations in s-Aff-Wild2}
\label{va_annot}
\end{figure}

Figure \ref{va_annot} illustrates the 2D Valence-Arousal histogram of annotations in s-Aff-Wild2.
Table \ref{expr_distr} displays the distribution of the eight expression annotations in s-Aff-Wild2.
Table \ref{au_distr} lists the twelve annotated AUs, their corresponding actions, and their distribution in s-Aff-Wild2.

\begin{table}[!h]
\caption{Multi-Task-Learning Challenge:  Number of Annotated Images for each of the 8 Expressions}
\label{expr_distr}
\centering
\begin{tabular}{ |c||c| }
\hline
 Expressions & No of Images \\
\hline
\hline
Neutral & 37,073  \\
 \hline
Anger & 8,094  \\
 \hline
Disgust & 5,922 \\
 \hline
Fear &  6,899 \\
 \hline
Happiness & 32,397  \\
 \hline
Sadness & 13,447  \\
 \hline
Surprise & 9,873  \\
 \hline
 Other & 39,701 \\
 \hline
\end{tabular}
\end{table}

\begin{table}[!h]
    \centering
        \caption{Multi-Task-Learning Challenge: Distribution of AU Annotations in Aff-Wild2}
    \label{au_distr}
\begin{tabular}{|c|c|c|}
\hline
  Action Unit \# & Action   &\begin{tabular}{@{}c@{}} Total Number \\  of Activated AUs \end{tabular} \\   \hline    
    \hline    
   AU 1 & inner brow raiser   & 29,995 \\   \hline    
   AU 2 & outer brow raiser  & 14,183 \\   \hline   
   AU 4 & brow lowerer   & 31,926 \\  \hline    
   AU 6 & cheek raiser  & 49,413 \\  \hline    
   AU 7 & lid tightener  & 72,806 \\  \hline    
   AU 10 & upper lip raiser  & 68,090 \\  \hline    
   AU 12 & lip corner puller  & 47,820 \\  \hline    
   AU 15 & lip corner depressor  & 5,105 \\  \hline   
  AU 23 & lip tightener & 6,538 \\  \hline    
   AU 24 & lip pressor & 8,052 \\  \hline    
   AU 25 & lips part  & 122,518 \\  \hline     
   AU 26 & jaw drop  & 19,439 \\  \hline     
\end{tabular}
\end{table}

The s-Aff-Wild2 dataset is divided into training, validation, and test sets. Initially, the training and validation sets, along with their annotations, are made available to participants for developing and testing their methodologies. The test set, however, is provided without annotations at a later stage.

Participants receive two versions of s-Aff-Wild2: cropped one and cropped-aligned one. All images/frames are processed using RetinaFace \cite{deng2020retinaface} to extract face bounding boxes and five facial landmarks. The images are cropped based on these bounding boxes. All cropped images have dimensions of $112 \times 112 \times 3$. The cropped-aligned version is obtained by applying similarity transformation using the five facial landmarks (two eyes, nose, and two mouth corners). Both versions are provided to participants, and the cropped-aligned version was used in our baseline experiments (described in Section \ref{baseline}).

\subsection{Compound Expression Recognition Challenge}

This Challenge utilizes a subset of the C-EXPR-DB database, consisting of 56 videos. The full C-EXPR-DB is an audiovisual in-the-wild dataset with 400 videos totaling approximately 200,000 frames, each annotated with 12 compound expressions. For this Challenge, the focus will be on recognizing the following seven compound expressions: Fearfully Surprised, Happily Surprised, Sadly Surprised, Disgustedly Surprised, Angrily Surprised, Sadly Fearful, and Sadly Angry.

Participants receive an unannotated subset of C-EXPR-DB (56 videos with about 26,500 frames) and must develop methodologies to recognize the seven specified compound expressions on a per-frame basis. Approaches may include supervised or self-supervised learning, domain adaptation, and zero- or few-shot learning techniques.

\section{Evaluation Metrics for each Challenge}\label{metrics}

In this subsection, we detail the metrics used to evaluate the performance of methodologies developed by participating teams in each challenge.

\subsection{Multi-Task Learning Challenge} 

The performance assessment for this Challenge is a composite measure that includes the following:

\begin{itemize}
    \item Valence and Arousal Estimation: Evaluated using the Concordance Correlation Coefficient (CCC).
\item Expression Recognition: Evaluated using the macro F1 Score across eight expression categories.
\item Action Unit Detection: Evaluated using the binary F1 Score for each of the twelve AUs.
\end{itemize}

Concordance Correlation Coefficient (CCC):
CCC measures the agreement between two variables and ranges from -1 to 1, with higher values indicating better agreement. It is calculated as follows:

\begin{equation} \label{ccc}
\rho_c = \frac{2 s_{xy}}{s_x^2 + s_y^2 + (\bar{x} - \bar{y})^2},
\end{equation}

\noindent
where $s_x$ and $s_y$ are the variances of all video valence/arousal annotations and predicted values, respectively, $\bar{x}$ and $\bar{y}$ are their corresponding mean values and $s_{xy}$ is the corresponding covariance value.

F1 Score:
F1 Score is the harmonic mean of precision and recall, providing a balance between the two. It ranges from 0 to 1, with higher values indicating better performance. It is defined as follows:

\begin{equation} \label{f1}
F_1 = \frac{2 \times precision \times recall}{precision + recall}
\end{equation}

Overall Evaluation Criterion:
The total performance measure for the Multi-Task Learning Challenge is the sum of the average CCC for valence and arousal, the macro F1 Score for expressions, and the binary F1 Score for AUs. The formula is as follows:

\begin{align} \label{mtll}
\mathcal{P}_{MTL} &= \mathcal{P}_{VA} + \mathcal{P}_{EXPR} + \mathcal{P}_{AU} \nonumber \\
&=  \frac{\rho_a + \rho_v}{2} + \frac{\sum_{expr} F_1^{expr}}{8} + \frac{\sum_{au} F_1^{au}}{12}
\end{align}

\subsection{Compound Expression Recognition Challenge}\label{evaluation4}

The evaluation metric for this challenge is based on the average F1 Score across the seven compound expressions. Thus, the performance criterion for the Compound Expression Recognition Challenge is defined as follows:

\begin{equation} \label{ce}
\mathcal{P}_{CE} = \frac{\sum_{expr} F_1^{expr}}{7}
\end{equation}

This formula calculates the mean F1 Score for all seven compound expressions, ensuring a balanced evaluation of performance across the different expression categories.

\section{Baseline Network and Performance} \label{baseline}

The baseline system is built entirely using open-source machine learning libraries to guarantee the reproducibility of results. Implemented in TensorFlow, the training process took approximately five hours on a Titan X GPU, utilizing a learning rate of  $10^{-4}$ and a batch size of 128.

\subsection{Multi-Task Learning Challenge}

The baseline model utilized is a VGG16 network with its convolutional layers fixed, training only the three fully connected layers. This model was pre-trained on the VGGFACE dataset. The output layer includes 22 units: 2 linear units for valence and arousal predictions, 8 softmax units for expression predictions, and 12 sigmoid units for action unit predictions. During training with the cropped-aligned version of the s-Aff-Wild2 database, the MixAugment technique \cite{psaroudakis2022mixaugment} was employed for data augmentation. Additionally, all image pixel intensity values were normalized to the range $[-1,1]$.

Table \ref{mtl} presents the evaluation results on the s-Aff-Wild2 validation set, showing the combined score of the average CCC for valence and arousal, the average F1 score for the expression categories, and the average F1 score for the AUs.

\begin{table*}[h!]
\caption{Multi-Task Learning Challenge results of  the baseline model on the validation set of s-Aff-Wild2; overall metric is in \%} 
\label{mtl}
\centering
\scalebox{1.}{
\begin{tabular}{ |c||c|c| }
 \hline
\multicolumn{1}{|c||}{\begin{tabular}{@{}c@{}} Teams \end{tabular}} & 
\multicolumn{1}{c|}{\begin{tabular}{@{}c@{}} Overall Metric  \end{tabular}} &
\multicolumn{1}{c|}{\begin{tabular}{@{}c@{}} Github \end{tabular}}  \\ 
  \hline
 \hline

VGGFACE with MixAugment  \cite{psaroudakis2022mixaugment} 
&  32.0 &   - \\
VGGFACE without MixAugment \cite{psaroudakis2022mixaugment} 
&  30.0 &   - \\
\hline
\end{tabular}
}
\end{table*}

\section{Conclusion}

In this paper we have presented the seventh Affective Behavior Analysis in-the-wild Competition (ABAW)  held in conjunction with ECCV 2024. This Competition follows its predecessors from CVPR 2024, 2023, 2022 \& 2017, ECCV 2022, ICCV 2021 and IEEE FG 2020. It features two challenges: Multi-Task Learning (MTL) to concurrently address Valence-Arousal Estimation, Expression Recognition, and Action Unit Detection; and Compound Expression Recognition, focusing on recognizing seven compound expressions. The databases utilized for this Competition are s-Aff-Wild2 and a subset of C-EXPR-DB.

\clearpage
\par\vfill\par

\bibliographystyle{splncs04}
\bibliography{main}
\end{document}